\documentclass[letterpaper, 10pt, conference]{ieeeconf}

\IEEEoverridecommandlockouts                              

\usepackage{dsfont}
\usepackage{lipsum}
\usepackage{amsmath}
\usepackage{graphicx}                       
\usepackage{stfloats}
\usepackage[tight,footnotesize]{subfigure}  

\usepackage{amssymb,amsmath}
\usepackage{textcomp}
\usepackage{mdwmath}
\usepackage{mdwtab}
\usepackage{eqparbox}
\usepackage{bm}
\usepackage{rotating}                       
\usepackage{array}                          
\usepackage{listings}                       
\usepackage[ruled, vlined, linesnumbered]{algorithm2e}
\usepackage{listings}
\usepackage[noend]{algpseudocode}
\usepackage{CJK}
\usepackage{indentfirst}
\usepackage{amsmath}
\usepackage{cite}
\makeatletter
\let\NAT@parse\undefined
\makeatother
\usepackage[colorlinks,
            linkcolor=blue,
            anchorcolor=blue,
            citecolor=blue,
            bookmarks=true
            ]{hyperref}
\usepackage{cleveref}

\usepackage{graphicx}
\usepackage{float}
\usepackage{subfigure}
\usepackage{amsmath}
\usepackage{algpseudocode}
\usepackage{stfloats}
\usepackage{color}

\usepackage{multirow} 
\usepackage{xcolor}
\usepackage[ruled,vlined]{algorithm2e}
\usepackage{amssymb}
\usepackage{booktabs}
\title{\LARGE \bf Real-Time Robot Navigation and Manipulation with Distilled Vision-Language Models}
\author{Kangcheng Liu,~\IEEEmembership{Member, IEEE}%
 }
\vspace{-1.01mm}

\begin{document}

\maketitle
\thispagestyle{empty}
\pagestyle{empty}

\vspace{-1.01mm}
\begin{abstract}
Autonomous robot navigation within the dynamic unknown environment is of crucial significance for mobile robotic applications including robot navigation in last-mile delivery and robot-enabled automated supplies in industrial and hospital delivery applications. Current solutions still suffer from limitations, such as the robot cannot recognize unknown objects in real-time and cannot navigate freely in a dynamic, narrow, and complex environment. We propose a complete software framework for autonomous robot perception and navigation within very dense obstacles and dense human crowds. First, we propose a framework that accurately detects and segments open-world object categories in a zero-shot manner, which overcomes the over-segmentation limitation of the current SAM model. Second, we proposed the distillation strategy to distill the knowledge to segment the free space of the walkway for robot navigation without the label. In the meantime, we design the trimming strategy that works collaboratively with distillation to enable lightweight inference to deploy the neural network on edge devices such as NVIDIA-TX2 or Xavier NX during autonomous navigation. Integrated into the robot navigation system, extensive experiments demonstrate that our proposed framework has achieved superior performance in terms of both accuracy and efficiency in robot scene perception and autonomous robot navigation. 
\vspace{-0.22mm}
\end{abstract}

\vspace{-0.083cm}
\section{Introduction}
\vspace{-0.113cm}
\vspace{-0.00666cm}
Robot scene perception and navigation are of essential significance to the development of human-like robot perception and navigation systems~\cite{liu2023fac, liu2023dlc, liu2023rm3d, liu2020fg, liu2022industrial, liu2022weakly, liu2022ws3d, hong2024liv, liu2024online, liu2022robust}.
With the development of large-scale vision-language models such as CLIP~\cite{radford2021learning} and SAM~\cite{kirillov2023segment}, the machine vision system gradually has the capacity to recognize the novel unseen object categories beyond the training set. However, several major limitations hinder the further application of large-scale pre-trained vision-language models (VLM) from wide deployments in a myriad of real-world robot applications such as navigation and grasping. The first is the \textit{computational resource limitations}. VLMs are computationally intensive and require significant processing power to perform tasks efficiently. For example, CLIP~\cite{radford2021learning} contains 63 million parameters merely for language transformer, and the Segment-Anything-Model (SAM)~\cite{kirillov2023segment} with ViT-H has 636 million parameters while GPT-3~\cite{dale2021gpt} has 175 billion parameters. Robots often have limited computational resources due to power constraints and size limitations. The second is \textit{information variability}. Real-world robot applications often encounter diverse and dynamic environments. Pre-trained models might not have encountered the wide range of scenarios that a robot could face, leading to performance degradation in novel situations. The third is \textit{limited labeled data}, which means fine-tuning these models for specific robot tasks might require labeled data, which can be scarce or extremely expensive to collect in real-world robotic navigation and inspection scenarios.

\begin{figure}[tbp!]
\setlength{\abovecaptionskip}{-0.35cm}
\setlength{\belowcaptionskip}{-0.38cm}
\centering
\includegraphics[width=\linewidth]{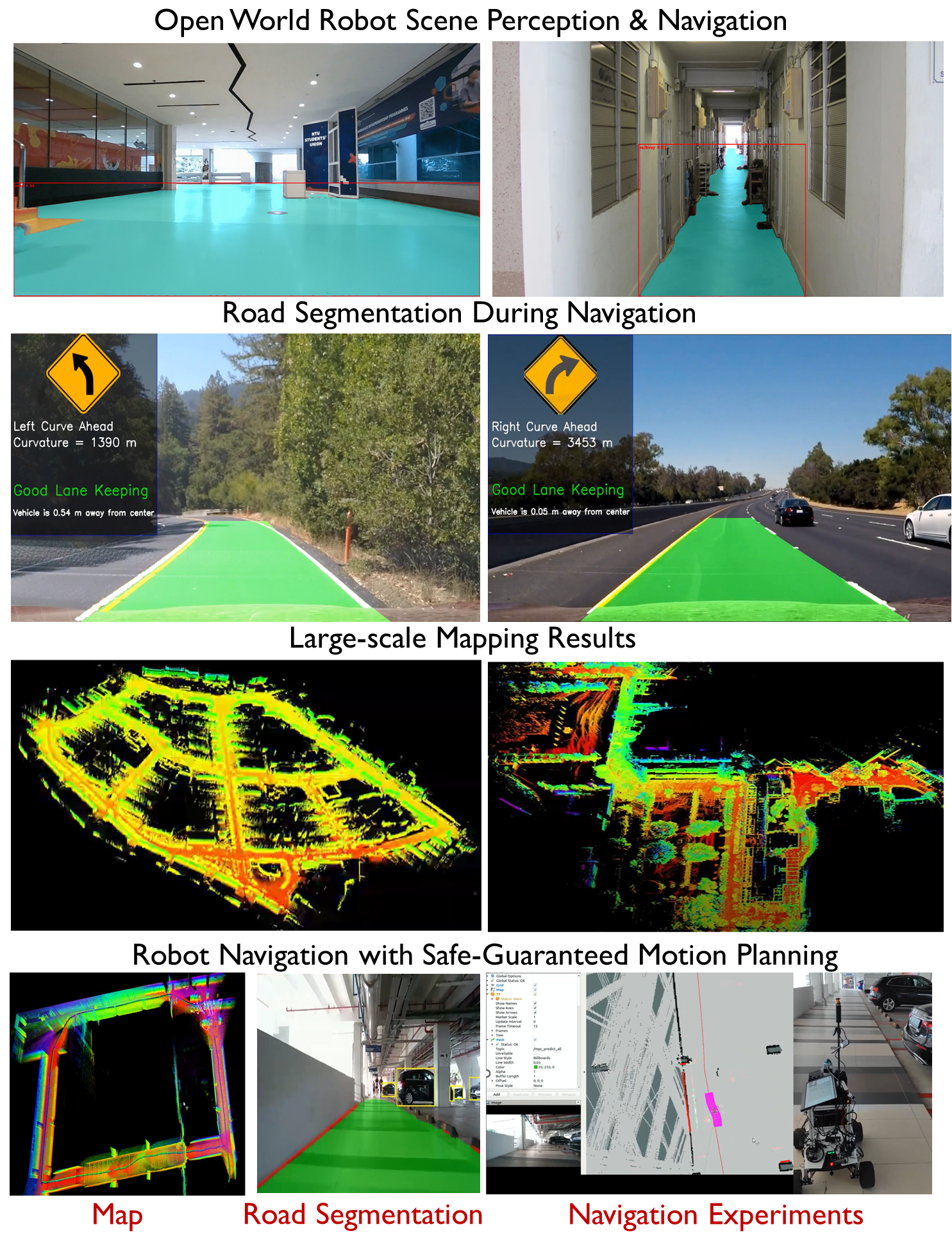}
\caption{Teaser: The autonomous navigation experiments in real-world situations. It can be demonstrated that our proposed approach can provide accurate segmentation results of the free space of the road and maintain real-time efficiency in the meantime. }
\label{robot_nav}
\end{figure}

\begin{figure*}[t]
\setlength{\abovecaptionskip}{-1.3mm}
\setlength{\belowcaptionskip}{-1mm}
\centering
\includegraphics[width=\linewidth]
{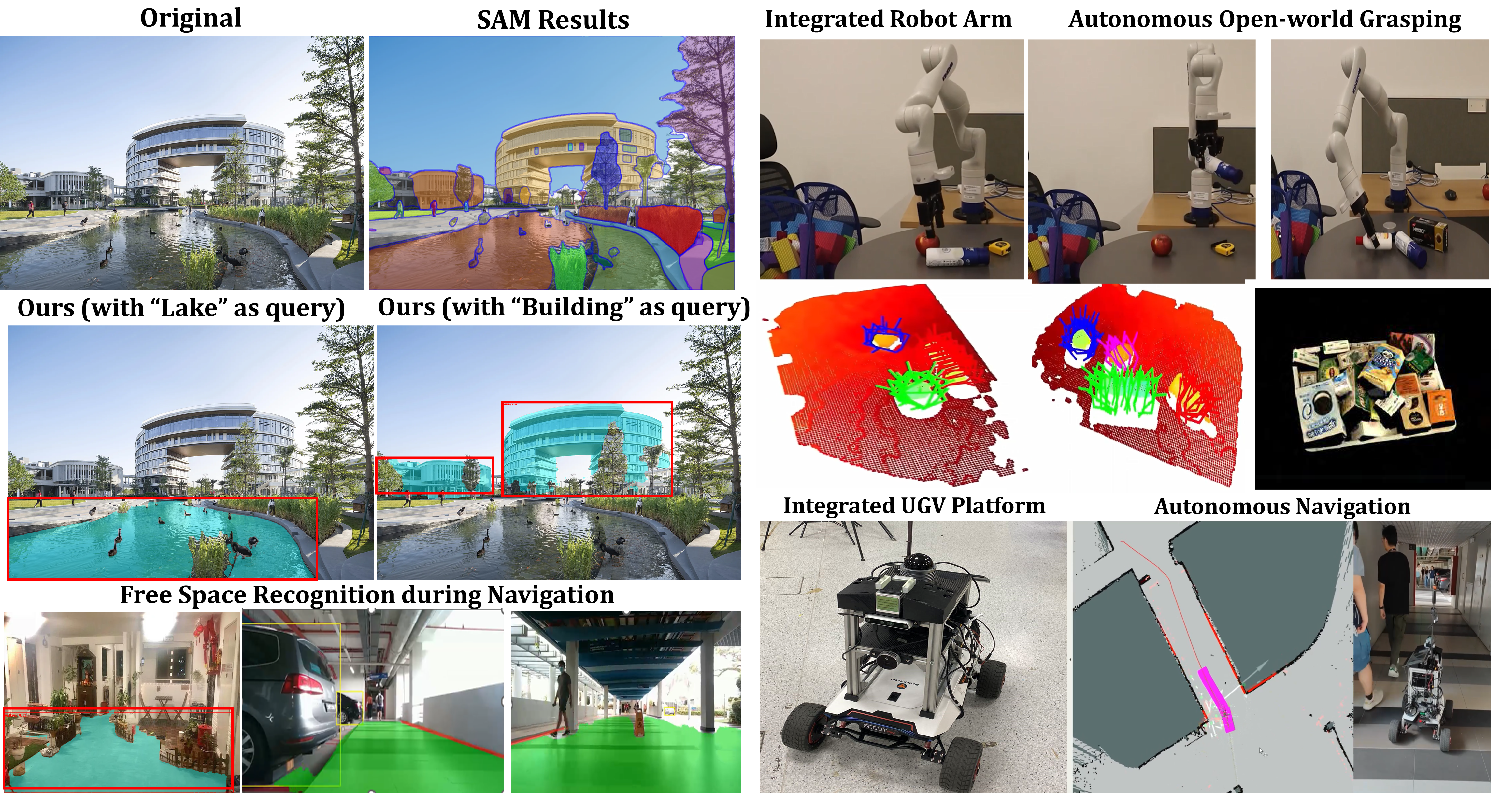}
\caption{ We have distilled the knowledge to a lightweight model that can run on the robot's onboard computer to help the robot navigate in real-world circumstances. Also, the robot grasping based on the open-vocabulary language prompted input can be realized. Segmentation comparisons in the open-world scenarios. Compared with the current prevailing SAM model, our proposed approach captures more holistic object semantic information.}
\label{Intro}
\end{figure*}


According to our experiments, when deployed into real-world applications, the current SAM~\cite{kirillov2023segment} suffers from over-segmentation and has trouble recognizing holistic object semantic information. The training dataset of SAM termed SA-1B contains more than 1 billion masks of 11M images. Although it excels at having a high level of scene parsing granularity, it might focus too much on capturing small geometric region-level superpixel details while overlooking semantic higher-level object representations. As demonstrated in Figure~\ref{Intro}, the deployed SAM model has poor performance in recognizing the holistic object and suffers from over-emphasizing the fine-grained information. Therefore, to endow the model with the open vocabulary recognition capacity to recognize novel objects, we designed an effective approach that effectively learns the vision-language aligned representation. To be more specific, we utilize language-aided semantic detection results to refine and rectify the over-segmented results provided by SAM. The detailed procedures are summarized as follows: 

Firstly, we propose the feature fusion and enhancement bi-directional transformer to enable more profound interaction between the vision and language modalities. Second, we propose a hierarchical grounding and alignment strategy to ground the specific phrase to the corresponding image regions provided by the region proposal network. Most importantly, we propose leveraging the CLIP~\cite{armeni20163d} with a detection head to refine the SAM~\cite{kirillov2023segment} segmentation with more holistic object-level information. Integrating the above components into a whole framework, we deployed the proposed approach to benefit autonomous robot navigation. As shown in Figure~\ref{Intro}, which exhibits a campus scene in the real-world environment, SAM~\cite{kirillov2023segment} suffers from over-segmenting objects into separated parts, while our proposed approach can maintain precise holistic object-level information, which is of great significance to general holistic visual recognition with accurate semantics, and our subsequent free space and speed limiter segmentation for visual recognition task.  


However, merely endowing the model with open-vocabulary recognition capacity cannot guarantee the real-time deployment of the vision-language models. To this end, we propose the distilling strategy to distill the knowledge of the vision language model to a lightweight network to achieve real-time performance in robotic navigation applications. Specifically, for the road recognition task in the campus environment. Moreover, we designed pruning strategies that can significantly reduce the size and complexity of a neural network, which can lead to significant savings in terms of both memory and computational resources. Extensive experiments demonstrate the effectiveness and efficiency of our proposed approach in boosting the real-world applications of robot autonomous navigation and vision-language navigation. Moreover, we integrate the proposed recognition system with the localization and mapping module as well as the motion planning module to achieve fully autonomous navigation in different human-dense complex environments.   

Here, we summarize several prominent contributions of our proposed framework:
\begin{enumerate}


\item  We proposed an effective image-level region-language matching approach, which effectively overcomes over-segmentation problems of the prevailing SAM~\cite{kirillov2023segment} model and provides a very powerful vision-language representation with accurate regional feature embedding alignment. Therefore, both the recognition accuracy on diverse public benchmarks and for real-world experiments are significantly improved. Also, it enables the recognition of unseen novel categories in the real-world environment in a zero-shot unsupervised manner.   \\
\vspace{-3.6mm}
\item We designed a lightweight distillation for distilling the knowledge from large-scale language models for conducting universal open-vocabulary segmentation in real-world scenarios. Meanwhile, we propose an effective trimming approach that works collaboratively with distillation to make the network lightweight to achieve real-time performance.




\item  By integrating with our proposed LiDAR-Inertial SLAM system and motion planning approaches with traversability mapping~\cite{liu2023dlc}, our framework can enable fully autonomous navigation within the dense dynamic pedestrian crowd. The designed framework is versatile and can support fully autonomous goal-point navigation with user-specified commands to tackle different complex situations in diverse real-world navigation and scene perception applications. 
\end{enumerate}



\section{Related Work}
\vspace{-0.19mm}
\vspace{-0.002mm}

\subsection{Open-Vocabulary Recognition for Robust Navigation}
\vspace{-0.002mm}
  The robot navigation has witnessed remarkable progress in recent decades~\cite{liu2022d}. However, the recognition large relies on the closed-set assumption, which means the robot merely has the capacity to recognize the object categories that appear within the training set~\cite{liu2023fac}. In real robotic applications such as exploration within the unknown environment and robot grasping and manipulations of novel objects, the recognition capacity of novel classes present in the environment is essential. The recent development of vision-language foundation models such as CLIP~\cite{radford2021learning} and Flamingo~\cite{awadalla2023openflamingo} has endowed the deep learning models with the capacity to recognize unseen novel categories because they have learned explicit vision-language feature associations in the shared feature space. The feature representation is learned from a large number of image-text pairs which are crawled from the Internet in an unsupervised manner. The learned feature representations will be beneficial for diverse downstream tasks such as detection and segmentation. Although tremendous progress has been made in these fields, the model still suffers from over-segmentation or inaccurate segmentation. In this work, we propose an open-vocabulary recognition approach that can effectively facilitate regional vision-language alignment. It tackles semantic object detection and segmentation in an open-world manner without any labeled training samples.  

  
\vspace{-1.6666mm}

\subsection{Trimming Strategy for Network Acceleration}
\vspace{-0.3mm}
Network trimming is an emerging and increasingly important research area that has attracted increasing attention recently. It has a large potential to reduce the redundancy within the network and make the network lightweight enough for real-time deployment in robotic applications. The network trimming approaches~\cite{shen2022prune} can be roughly categorized into unstructured~\cite{tanaka2020pruning, kwon2020structured}, semi-structured~\cite{sui2023hardware, balasubramaniam2023r}, and structured trimming~\cite{fang2023depgraph, zhang2021structadmm} approaches. Structural pruning has its unique merits of easy and universal characteristics for deployment on different hardware and software without any modifications~\cite{liang2021pruning}. In this work, we aim to design a simple but effective structural pruning approach that works effectively for walkway free-space recognition, which facilitates subsequent autonomous robot navigation.

\vspace{-0.002 mm}

\begin{figure*}[tbp!]
\setlength{\abovecaptionskip}{-0.38cm}
\setlength{\belowcaptionskip}{-0.22cm}
\centering
\includegraphics[width=\linewidth]{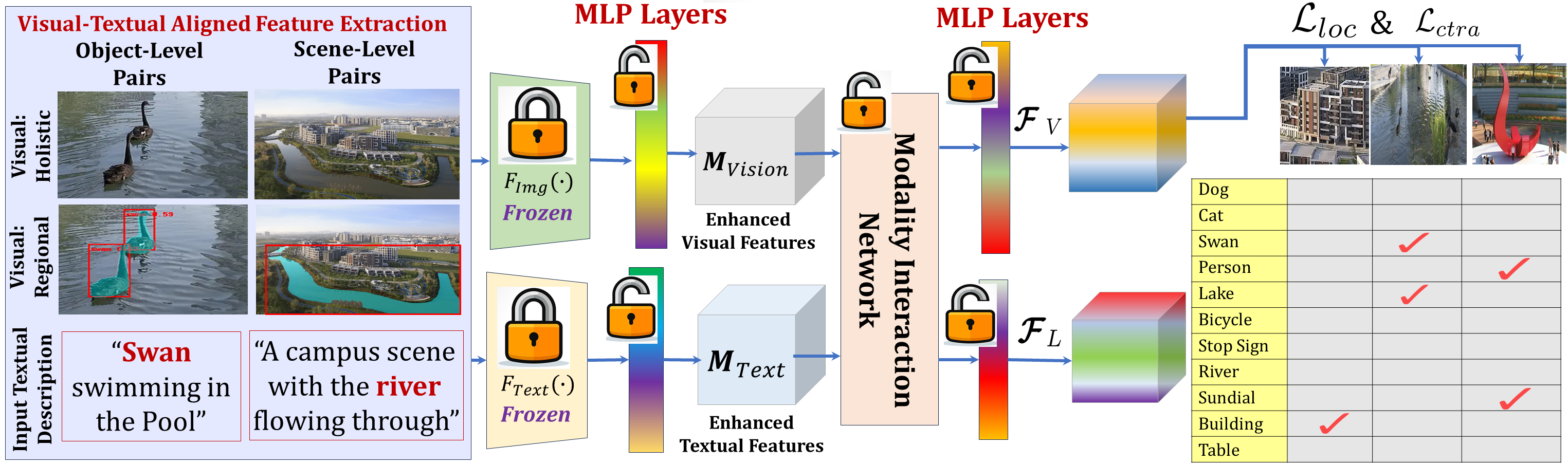}
\caption{The open-vocabulary detection results in real-world complicated scenes. Previous vision-language models CLIP~\cite{radford2021learning} can merely deal with the task of image classification and can not tackle the detection and segmentation required in robotic applications. While the SAM~\cite{kirillov2023segment} model focuses too much on fine-grained details and suffers from over-segmentation. Our proposed approach captures object-level information by region proposals and facilitates precise visual-language association through regional contrastive representation learning, which allows precise vision-language association at the regional level. Moreover, we design a modality interaction network to explore relations between the visual and linguistic modality. Also, it boosts the fusion of vision and linguistic features. According to our experiments on both public benchmarks and real-world experiments, these designs demonstrate superior open-vocabulary recognition accuracy and lead to successful autonomous robot navigation in real-world complex scenarios. }
\label{Figure_framework}
\vspace{-1.9 mm}
\end{figure*}


\begin{figure}[tbp!]
\setlength{\abovecaptionskip}{-0.1cm}
\setlength{\belowcaptionskip}{-0.1cm}
\centering
\includegraphics[width=\linewidth]{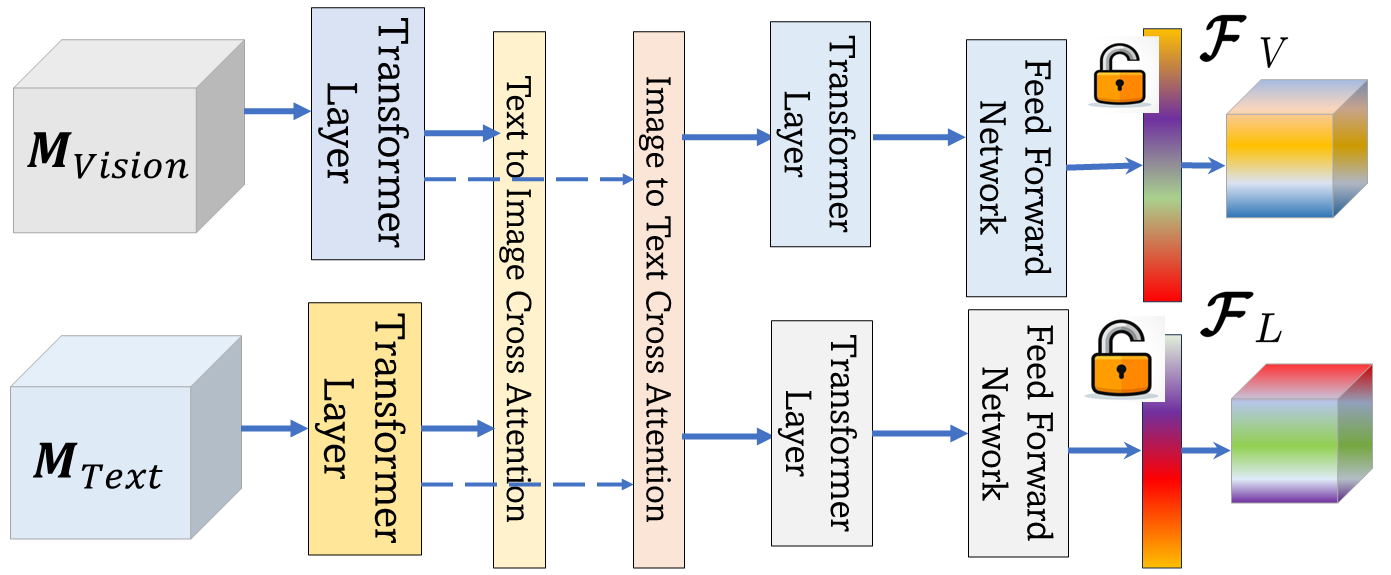}
\caption{The detailed structure of the proposed modality interaction Transformer network. The proposed network is simple but effective in capturing as well as modeling the rich cross-modality feature relations and interactions within the vision and linguistic modality.}
\vspace{-0.3 mm}
\label{Figure_att_modules}
\end{figure}

\vspace{-1mm}
\subsection{Knowledge Distillation for Robotic Perception Tasks and Vision-Language Model-Enabled Robot Navigation}
\vspace{-1mm}
Knowledge distillation has been demonstrated as a very effective approach for transferring knowledge from the original large-sized model with rich knowledge to the small-sized model for lightweight image processing tasks~\cite{guo2023class}. We propose an effective knowledge distillation approach to conduct knowledge transferred from the original vision-language models.
Although robot localization and mapping techniques have witnessed tremendous progress in the past few years~\cite{liu2023dlc, li2021towards, liu2022enhanced, liu2022light, liu2022integrated, liu2020fg, liu2019deep}, the navigation system that can leverage the reliable perception capacity from the vision-language models remains in its infancy. Therefore, we take the first step in proposing the vision-language model to benefit robot scene perception as well as navigation tasks. We first propose an approach to enable faithfully inheriting the vision-language associated knowledge. 


\vspace{-0.000001cm}



\section{The Proposed Open-Vocabulary Recognition Approach}
In this section, we elaborate on our proposed approach to achieve open-vocabulary scene parsing. \textit{To start with}, we introduce our regional word-object matching approach, which establishes an explicit association between the vision and language information at both the fine-grained region level and the coarse-grained image level with the generated region proposals in an unsupervised manner. \textit{Second}, we designed the multi-modal feature interaction modeling with cross-modality transformers, which significantly boosts the final recognition performance. As shown in Fig.~\ref{Figure_framework}, we keep the original CLIP~\cite{radford2021learning} model weights frozen to maintain the training efficiency and not deteriorate the original well-aligned vision-language representations in CLIP which is learned from a staggering amount of 400 million image-text pairs in the meantime. We add fully connected MLP layers before the modality interaction network to obtain the enhanced features $\textit{\textbf{M}}_{Vision}$ and $\textit{\textbf{M}}_{Text}$ for visual and textual representations respectively. According to our extensive experiments, the performance is also largely improved based on the original CLIP model. 

\vspace{-1.0 mm}

\subsection{Vision-Language Bi-Directional Transformer for Feature Fusion and Feature Interaction Modeling}
\vspace{-2mm}
In this section, we propose a direction attentional interaction modeling module, which explicitly finds and enhances the vision-language feature associations within the training of the network, while boosting the vision-language feature discrimination of the irrelevant or distinct features.
The fusing of visual and linguistic features can be formulated as the cross-attention interaction modeling operation with the visual feature as the anchor:

\vspace{-1.36mm}

\begin{equation}
\textit{\textbf{F}}_{V\rightarrow{L}}^{\bigstar}=softmax(\frac{(W^{m}_qV)^T (W_k^{m}T)}{\sqrt{\hat{N}}}) (W_v^mT)^T,
\label{att_v_l}
\vspace{-3mm}
\end{equation}
\\
Where $W^{m}_q$, $W^{m}_k$, and $W^{m}_v$ are the weights of the transformation functions that are implemented by the multi-layer perceptron (MLP). These weights help to unify the dimension of channels to $\hat{N}$. The operation can be interpreted as that we find the correlated visual feature with respect to each single word embedding adaptively with the training of the vision-language model as illustrated in Fig.~\ref{Figure_framework}. 

On the other hand, we conduct the correlation mining between the visual and linguistic features with the textual features as the anchor, which can be given as follows:
\vspace{-1mm}

\begin{equation}
\textit{\textbf{F}}_{L\rightarrow{V}}^{\bigstar}=softmax(\frac{(W^{m}_qE)^T (W_k^{m}V)}{\sqrt{\hat{N}}}) (W_v^mV)^T
\label{att_l_v}
\vspace{-1mm}
\end{equation}

The operation can be interpreted as that we find the correlated word-level feature with respect to each single pixel embedding adaptively with the training of the vision-language model as illustrated in Fig.~\ref{Figure_framework}. 



The simple self-attentional transformer can implicitly model the interaction between the language and visual features. As shown in Figure \ref{Figure_att_modules}, integrated with transformer layers as well as the final feed-forward network, the proposed modality interaction network can effectively model the vision-language correlations and interactions during the pre-training of the vision-language model. However, the above operation in Eq. \ref{att_l_v} and Eq. \ref{att_v_l} merely considers the relationship between a single image pixel as well as a single word. Also, the pixel-level bi-directional cross-modality interaction modeling is computationally very expensive, adding an extra computational burden during the pre-training stage. Therefore, we propose directly using the regional visual feature for effective interaction modeling. We conduct max-pooling operations based on the pixel-level feature $V$ to obtain the regional feature $V^R$, which is effective in retaining the most prominent feature within the region. Based on Eq. \ref{att_v_l} and Eq. \ref{att_l_v}, the final visual and textual texture features are given as $\textit{\textbf{F}}_{V^R\rightarrow{L}}^{\bigstar}$ and $\textit{\textbf{F}}_{L\rightarrow{V^R}}^{\bigstar}$. As shown in Fig. \ref{Figure_att_modules}, these features are fed through the transformer layers and feed-forward network to obtain the final feature $\mathcal{F}_L$ and  $\mathcal{F}_V$, which facilitates
subsequent vision-language-matched contrastive learning. 
\vspace{-1.0mm}
\subsection{Regional Vision-language Matching Strategy}


\begin{algorithm}[t!]
\footnotesize
     \caption{The Proposed Network Trimming and Distillation Approach.}     \label{alg_clustering}
      \KwIn{The training data and the original model} 
      \textbf{Given}: The target FLOPs decreasing rate ($\mathcal{R}$).
  \label{alg_1}  \\
  Initialize the network weight with the distillation model. \\
  Initialize the current pruning rate as 0\%;\\
 Random sampling the dataset $R^{'}$ from original dataset  
 $R$. \\
 Train the original model with distillation loss for $t_{start}$ epochs. 
 \\
  \While{$\mathcal{P}^{'}<\mathcal{P}$}
 {
 \For{l $\leftarrow$ 1 to   L}  {
          \For{c $\leftarrow$ 1 to c }{Calculate the importance of the $l_{th}$ filter utilizing the $c_{th}$ filter feature selection criteria; \\
          Remove the TOP-3 filter with the lowest importance scores $\mathcal{S}_{total}$;
          $W_{trimmed}$ $\leftarrow$ $W_{1}$; \\ 
        
          }
        }   
        Train the pruned model with the loss $\mathcal{L}_{dist}$ until convergence; \\
        Calculate current pruning rate $\mathcal{R}^{'}$.\\
        }
        \KwOut{The pruned model for deploying onto the robot platform.}
        \vspace{-0.7mm}
        \end{algorithm}


We propose regional contrastive learning to conduct the regional vision-language matching. Different from current prevailing approaches such as DINO-v2~\cite{oquab2023dinov2}, which conduct contrastive learning at the pixel level, we conduct the feature contrast at the region level. It can significantly improve the efficiency while the network is trained with large-scale image-text pairs.


The final contrastive loss consists of both the image-to-text contrastive loss and the text-to-image contrastive loss, the total loss is given as $\mathcal{L}_{Ctra}$ and is formulated as:
\begin{equation}
\mathcal{L}_{Ctra}=  \mathcal{L}^{I \rightarrow T}_{Ctra} +\mathcal{L}^{T \rightarrow I}_{Ctra}
\end{equation}
\vspace{-2.0666mm}

On the one hand, we have the image-to-text contrast loss $\mathcal{L}^{I\rightarrow T}_{Ctra}$, which facilitates the accurate matching between the image regional feature and the most precisely matched textual description can be formulated as follows:

\vspace{-2mm}

\begin{equation}
    \mathcal{L}^{I\rightarrow T}_{Ctra}=-\frac{1}{\mathcal{D}}\sum_{i=1}^{D} \log\frac{\exp(\bm{\mathcal{F}_V} \cdot \bm{\mathcal{F}}_{L} /\tau)}{\sum_{(\cdot, c) \in \textbf{\textit{B}}} \exp( \bm{\mathcal{F}}_V \cdot  \bm{\mathcal{F}}_L  /\tau))}.
\end{equation}


On the other hand, the text-to-image loss can be formulated as follows, which enables the precise matching between the textual and image-level features:
\vspace{-1mm}
\begin{equation}
    \mathcal{L}^{T \rightarrow I}_{Ctra}=-\frac{1}{\mathcal{D}}\sum_{i=1}^{D} \log\frac{\exp(\bm{\mathcal{F}_L} \cdot \textit{\bm{\mathcal{F}}}_{V} /\tau)}{\sum_{(\cdot, c) \in \textbf{\textit{B}}} \exp(\bm{\mathcal{F}}_L \cdot  \bm{\mathcal{F}}_V /\tau))}.
    \vspace{-3mm}
    \vspace{-3mm}
\end{equation}
\\

The most important component in vision language association is to find the matched contrast pairs. To this end, we propose a matching strategy based on feature-level similarity. We calculate the similarity of each element between the visual feature embedding and the language feature embedding. 
The vision language similarity can be given as follows:


\begin{small}
\begin{equation}
     \textit{S}_{V, L}=  \mathop{\arg\max}_{K} \mathcal{F}_{Sim}(\bm{\mathcal{F}}_L, \bm{\mathcal{F}}_{V}).
\end{equation}
\end{small}
\\   
Where $\mathcal{F}_{Sim}$ is the similarity function and we choose the cosine similarity to evaluate the degree of correlation between the visual and linguistic feature embeddings. We choose $K=3$, thereby finding the three most correlated region visual features with textual description and vice versa. The top-$K$ problem can be formulated as an optimal transport problem. Finally, this vision-language correlation can be beneficial to learning meaningful and informative representation. Also, by contrastive learning which pulls similar positive pairs together while pushing the different negative pairs apart, the discriminative representation in the vision-language co-embedded feature space can be effectively modeled and learned. Because we also have object detection with the region proposal network. The final pertaining training loss is formulated as $\mathcal{L}_{pretrain}= \mathcal{L}_{loc}+ \mathcal{L}_{ctra}$. The localization loss $\mathcal{L}_{loc}$ is adopted as the focal loss which focuses on learning better representations with an emphasis on hard misclassified examples. Meanwhile, we utilize the $L_2$ loss as well as the generalized IoU loss for conducting bounding box regressions. During the inference stage, we directly utilize the semantics provided by CLIP for the most pixels to merge the over-segmented regions provided by SAM~\cite{kirillov2023segment} and offer the final semantic predictions based on the similarities with the CLIP language query in the semantic vocabulary space for the whole region. As validated by our extensive real-world experiments, the learned embeddings can achieve very precise open-world learning and facilitate accurate free road space recognition which contributes to robot autonomous perception and navigation.  


\section{Proposed Distillation, Trimming, and Network Acceleration Approaches}
\vspace{-0.7mm}

The core idea within the structural trimming is to remove the comparatively insignificant filters. Hence, the key in structural pruning is to evaluate the importance of diverse filters. We propose methods to evaluate the significance of diverse filters in performing recognition and eliminate the redundant or insignificant ones. \\
\textbf{Layer Weight Magnitude-based Importance Selection.} As demonstrated by previous works~\cite{liu2023generalized}, the network weights having a larger magnitude and norm value are regarded as the most important in the network activation calculation, thus playing a dominant role in the network decision. 
Representing the $p_{th}$ filter in the $q_{th}$ network layer as $\mathcal{C}_p \in \mathbb{R}^{1\times M_h}$, the corresponding $l_z$ norm of it can be given as: 
\begin{equation}
\|\mathcal{C}^{a/b}_q\|=(\sum_{i=1}^{M_h}\|c_i\|^z)^{\frac{1}{z}} 
\vspace{-1mm}
\end{equation}

In our implementation, we adopt both $l_1$ and $l_2$ norms for a more comprehensive evaluation of the importance of network weights. 

\noindent
\textbf{Structural Similarity-based Importance Selection.} The magnitude can merely evaluate the significance of different filters but can not avoid redundancies existing within different filters.
To quantitatively evaluate the correlation of the filters and remove the redundant ones, we also adopt the similarity criteria which allows to evaluate the contribution and the substitutability of different filters. We adopt both Euclidean similarity and Cosine similarity to determine whether the filter can be substituted by the other remaining filters. Denote the two compared filters as $\mathcal{C}_a$ and  $\mathcal{C}_b$, respectively, the Euclidean similarity can be calculated as follows:
\begin{equation}
    \mathcal{S}_{euc}(\mathcal{C}_a, \mathcal{C}_b) = \sqrt{\sum_{i=1}^{M_h}\|c_a-c_b\|^2}
\end{equation}
The Cosine similarity can be formulated as follows:
\begin{equation}
    \mathcal{S}_{cos}(\mathcal{C}_a, \mathcal{C}_b) = 1-\frac{\sum_{i=1}^{M_h}(c_a \cdot c_b)}{\|c_a\|\|c_b\|}
\end{equation}
The final score during the training can be given as the sum of the trimming criteria designed above. The score is given as $\mathcal{S}_{total} =\mathcal{C}_1^a + \mathcal{C}_1^b + \mathcal{C}_2^a + \mathcal{C}_2^b + \mathcal{S}_{euc}(\mathcal{C}_a, \mathcal{C}_b) + \mathcal{S}_{cos}(\mathcal{C}_a, \mathcal{C}_b).$ The score should be as high as possible and the filter pairs that result in small scores can be removed to make the network lightweight. In practice, we will remove the TOP-3 filters with the lowest importance scores. Finally, we integrated the designed criteria into the network training with the details given in Algorithm \ref{alg_1}.

\begin{figure}[tbp!]
    \centering
    \includegraphics[width=\linewidth]{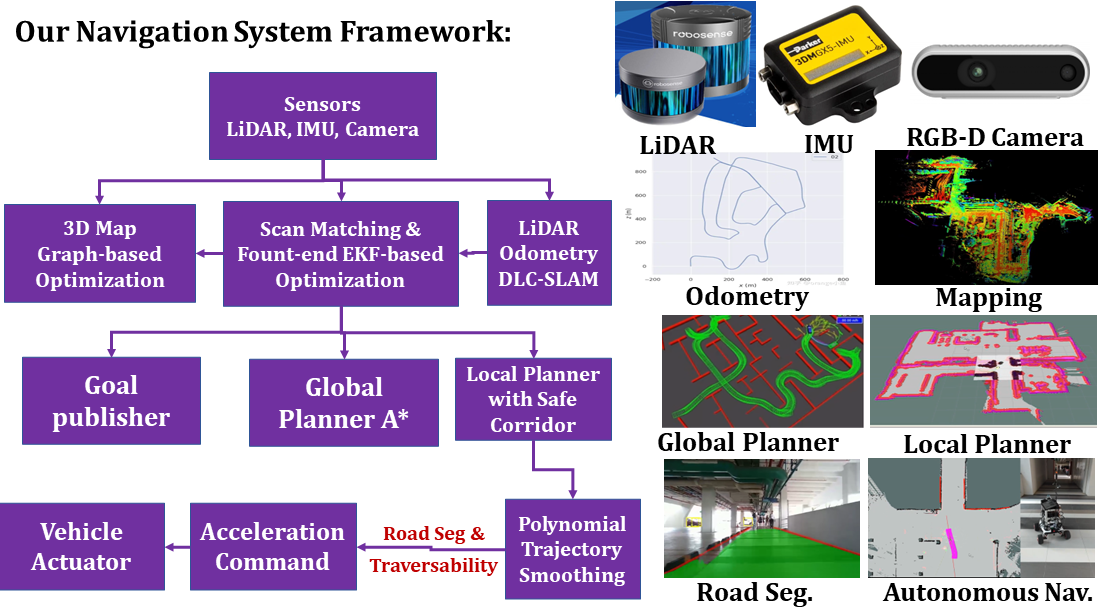}
    \caption{Our final integrated system framework achieves autonomous robot navigation in real-world environments. We first propose an open vocabulary recognition approach that recognizes unseen novel categories. Next, we distill the knowledge from the open-vocabulary model for free space recognition of the road, and proposed network trimming approaches to achieve real-time performance on the robot onboard computer. Integrated with the system framework depicted above which is extended from our previous work~\cite{liu2023dlc}, we perform autonomous language-guided navigation.}
    \vspace{-0.1mm}
    \label{Figure_system}

\end{figure}

\begin{figure}[t!]
\setlength{\abovecaptionskip}{-0.1cm}
\setlength{\belowcaptionskip}{-0.1cm}
\centering
\includegraphics[width=\linewidth]{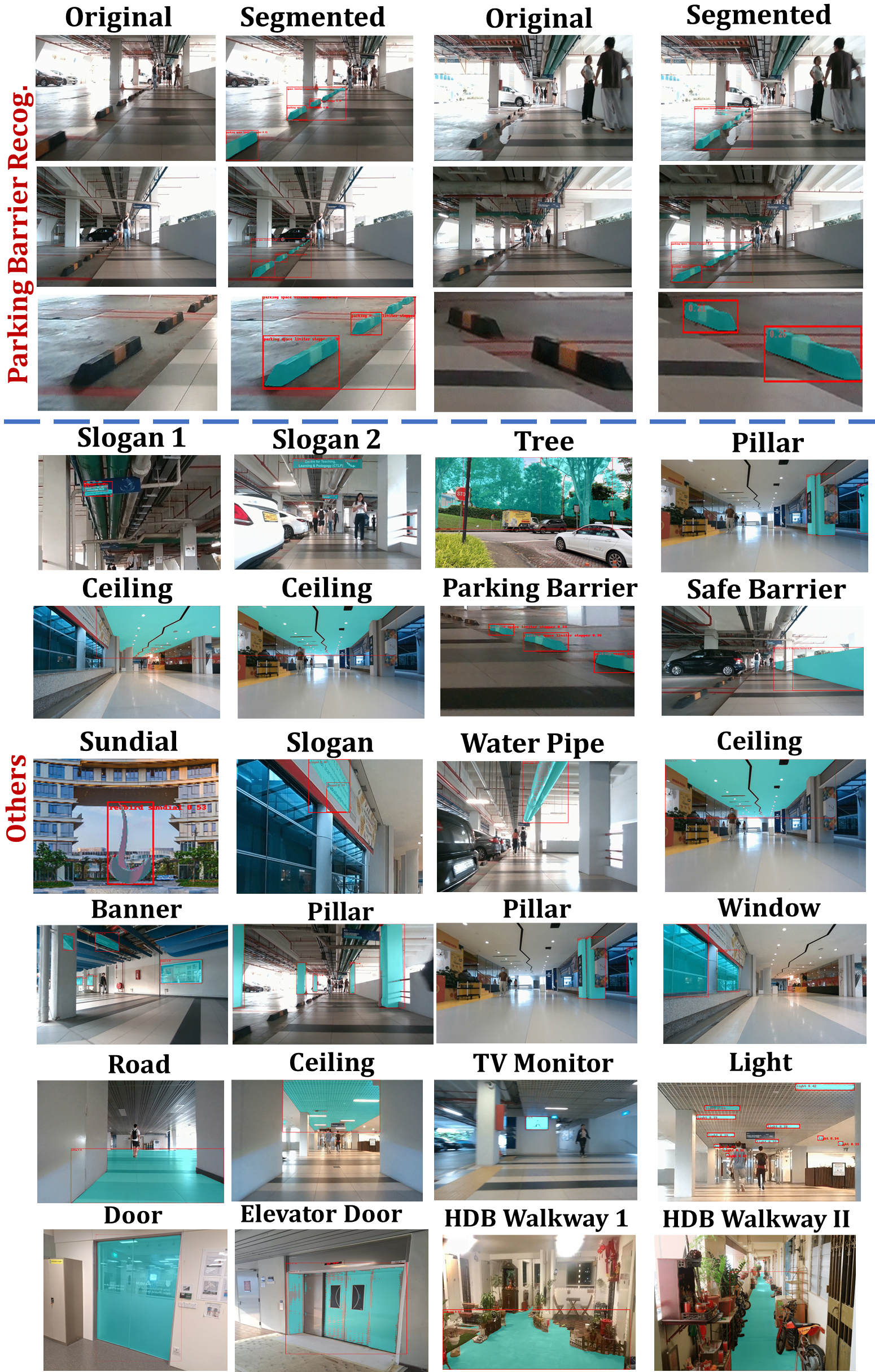}
\caption{The open-vocabulary detection results in real-world complicated scenes. It can be demonstrated that our proposed approach has satisfactory performance in complex environments in achieving open-world recognition.}
\label{Figure_real_world}
\end{figure}

\noindent
\textbf{The Knowledge Distillation-based Open-Vocabulary Recognition Capacity Transfer.}
On top of the designed regional vision language associated pre-training strategy. We also designed an effective knowledge-distillation strategy that can extract the knowledge for the large-scale vision language models to benefit the lightweight deployment in real robotic applications. We selected the lightweight network ResNet-50 as the student model to inherit the learned discriminative representations from the teacher model with balanced accuracy and efficiency. The 
designed distillation loss can be formulated as follows:
\vspace{-1mm}
\begin{equation}
    \mathcal{L}_{dist} =\frac{1}{N}\sum_i \mathcal{L}_{KL}(\mathcal{Y}_{VL}, \mathcal{Y}_{Light})
    \label{distill}
\end{equation}

For our circumstances, we are required to accurately segment the road and the curb line. Therefore, we directly utilize the word ``curb line" as the prompt for the language encoder in distilling the knowledge from the visual-linguistic models. The final experimental results demonstrate that our distillation strategy can successfully maintain the knowledge from the large-scale vision-language model, and by cooperating with our proposed network trimming approach in Algorithm \ref{alg_1}, real-time performance can be achieved in recognizing the free spaces of the road, which largely facilitates subsequent autonomous robot navigation.

\section{Experimental Results and System Integration for Autonomous Navigation}

\begin{table}[t]
    \caption{The benchmark testing with the inference video resolution of $1920 \times 1080$ on the YouTube-VIS~\cite{yang2019video} benchmark. Tested in an unsupervised manner, our proposed approach achieves better accuracy \& efficiency compared with fully supervised methods.}
    \label{table_youtube}
    \begin{center}
    \scalebox{0.59}{\begin{tabular}{lllllll}
    \toprule
    Method &Mask AP\% & Box AP\% & On 3090 & On 2080Ti & On Jetson Orin & On Xavier NX \\
    \hline
    \multicolumn{7}{l}{ Tested in a \textit{fully supervised} manner:} 
    \\
    \hline
    Efficient-Net + Det. Head & 44.7\% & 47.1\%  & 47.8 ms  & 88.82 ms  & 132.56 ms & 190.26 ms \\
    MobileNetV3 + Det. Head & 45.3\% & 48.2\%  & 42.9 ms  & 79.72 ms   & 118.98 ms & 170.77 ms \\  
    FCOS~\cite{tian2019fcos} & 45.9\% & 47.3\%  & 47.6 ms & 88.11 ms & 132.37 ms & 189.78 ms \\
    Faster-RCNN~\cite{ren2015faster} & 45.7\% &  47.5\% & 56.3 ms & 104.65 ms & 156.12 ms & 224.08 ms \\
    YOLO-ACT~\cite{yolactpami2022} & 46.2\% & 48.6\% & 56.7 ms & 105.37 ms & 157.29 ms & 225.76 ms\\
    YOLO-Edge~\cite{liang2022edge} & 48.8\%  & 51.2\%  & 54.6 ms & 101.52 ms & 151.39 ms & 217. 289 ms\\
    \hline
    \multicolumn{7}{l}{ Tested in a \textit{unsupervised} manner:} \\
    \hline
    Ours (Merely Distillation) &  62.5\% & 62.3\% & 39.1 ms & 72.61 ms & 108.55 ms & 155.80 ms \\
    Ours (After Trimming) & 55.8\% & 54.7\%  & 16.3 ms  & 30.31 ms & 45.28 ms &  64.99 ms \\
    \bottomrule
    \end{tabular}}
    \end{center}
\end{table}

\begin{figure}[t!]
\setlength{\abovecaptionskip}{-0.1cm}
\setlength{\belowcaptionskip}{-0.1cm}
\centering
\includegraphics[width=\linewidth]{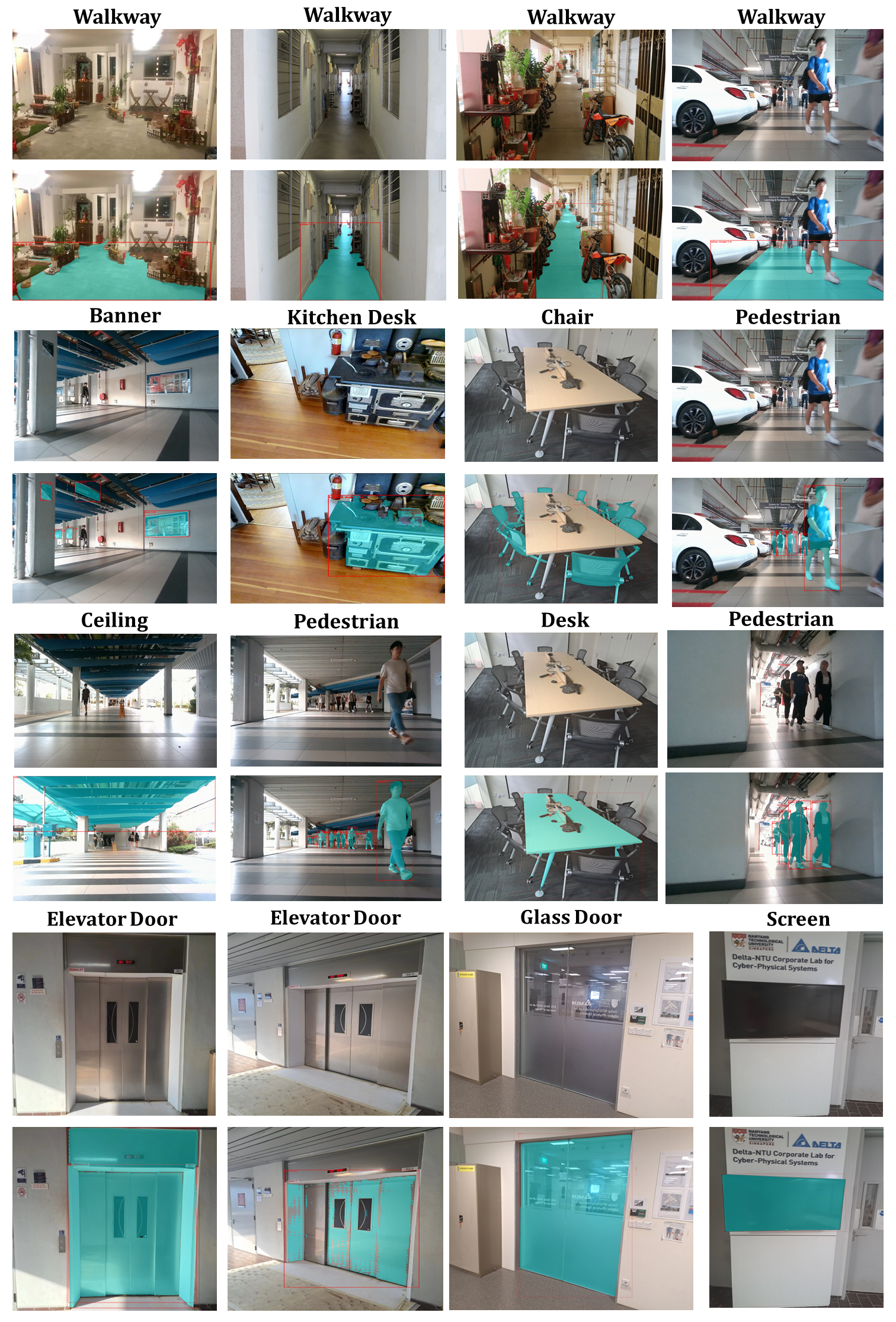}
\caption{More results of open-world recognition in complex environments.}
\label{Figure_real_world}
\vspace{-0.98mm}
\end{figure}

\begin{table}[t]
    \caption{The onboard inference speed on different devices in experimenting with Navigation ROSBag in the campus environment with the inference video resolution of $1280 \times 720$. The original model is not able to run on Nvidia-TX2 due to large GPU memory consumption. }
    \label{table_ros}
    \begin{center}
    \scalebox{0.71}{\begin{tabular}{llcccc}
    \toprule
     Method & On 3090 & On 2080Ti & On Jetson Orin & On Xavier NX & On TX2 \\
    \hline
    The Original Model & 126.9 ms &  235.7 ms  & 351.9 ms & 425.8 ms & N.A. \\
    After Trimming  & 16.8 ms   &   31.2 ms & 46.6 ms & 65.7 ms &  94.3 ms\\
    \bottomrule
    \end{tabular}}
    \end{center}
    \vspace{-0.98 mm}
\end{table}

\noindent
\vspace{-1.1 mm}
\subsection{Network Training Details}

For the pre-training of the vision-language models as shown in Fig.~\ref{Figure_framework}, we directly use the text-image pairs provided by the conceptual caption dataset CC3M~\cite{sharma2018conceptual}, which comprises three million text-image pairs on the Internet. The training was initialized from CLIP pre-trained weight with the backbone of the ViT-base-patch32. We used the Adam optimizer with a batch size of 80, an initial learning rate of 0.001, a maximum iteration of 300k, and 80 regions per image. The training merely lasts for 8.2 hours with four 2080Ti GPUs in parallel. As for the distillation process shown in Eq. \ref{distill}, we use the fine-tuned ViT-base-patch32 as the teacher model and the ResNet-50 as the student model. The training epoch $t_{start}$ is set as 100k to inherit the knowledge and well-aligned vision language representations from the large model (ViT-base) to a smaller model (ResNet50). The trimming is performed on the ResNet50 model with our designed approach in Algorithm \ref{alg_clustering}.



\begin{figure}[tbp!]
\setlength{\abovecaptionskip}{-0.21cm}
\centering
\includegraphics[width=\linewidth]{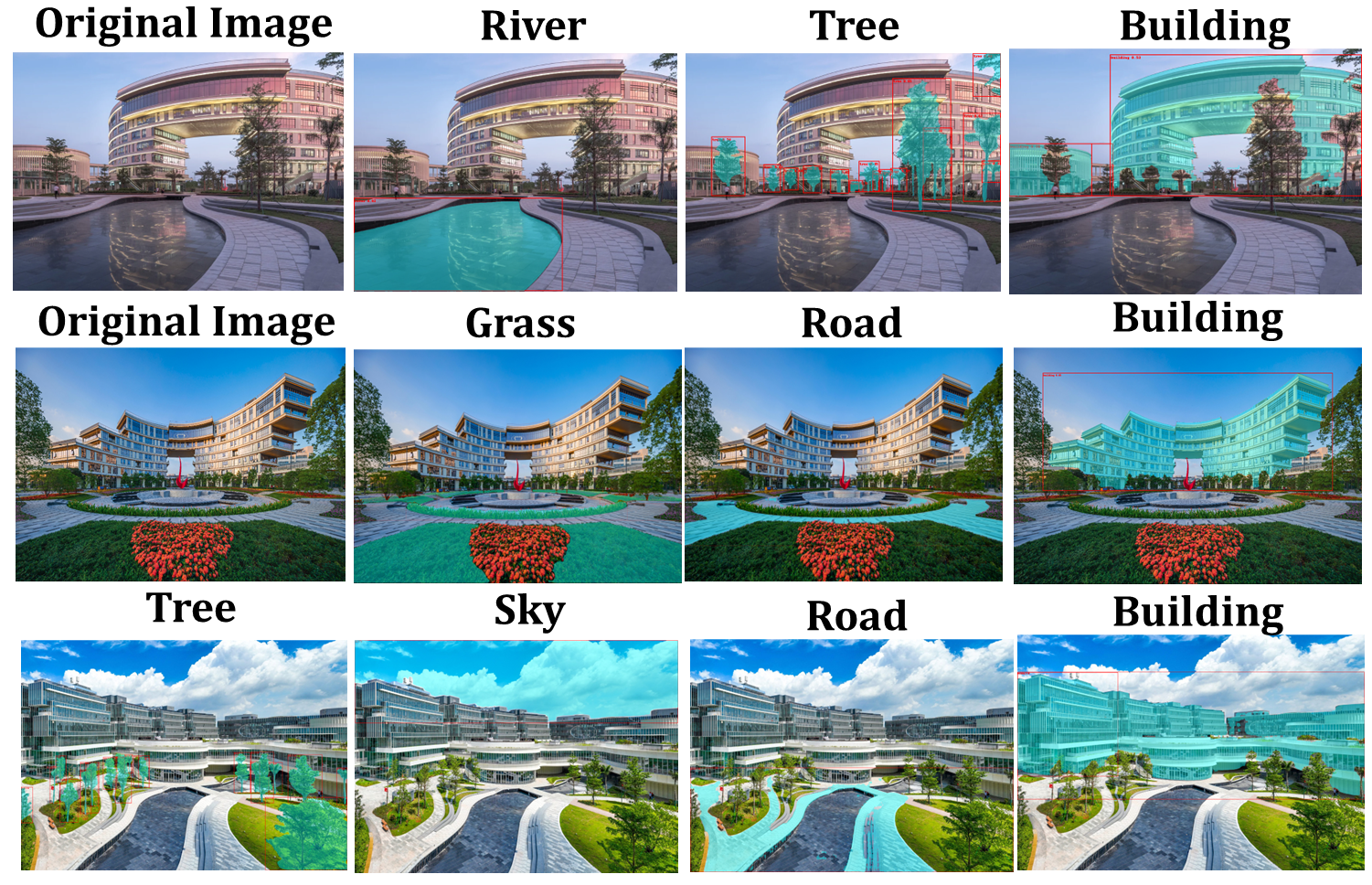}
\caption{Our outdoor open-vocabulary recognition experimental results. Diverse components of the outdoor scenes can be clearly separated by our proposed unsupervised open-vocabulary recognition approach.}
\label{building_seg}
\end{figure}

\begin{figure}[tbp!]
\setlength{\abovecaptionskip}{-0.36cm}
\setlength{\belowcaptionskip}{-0.59cm}
\centering
\includegraphics[width=\linewidth]{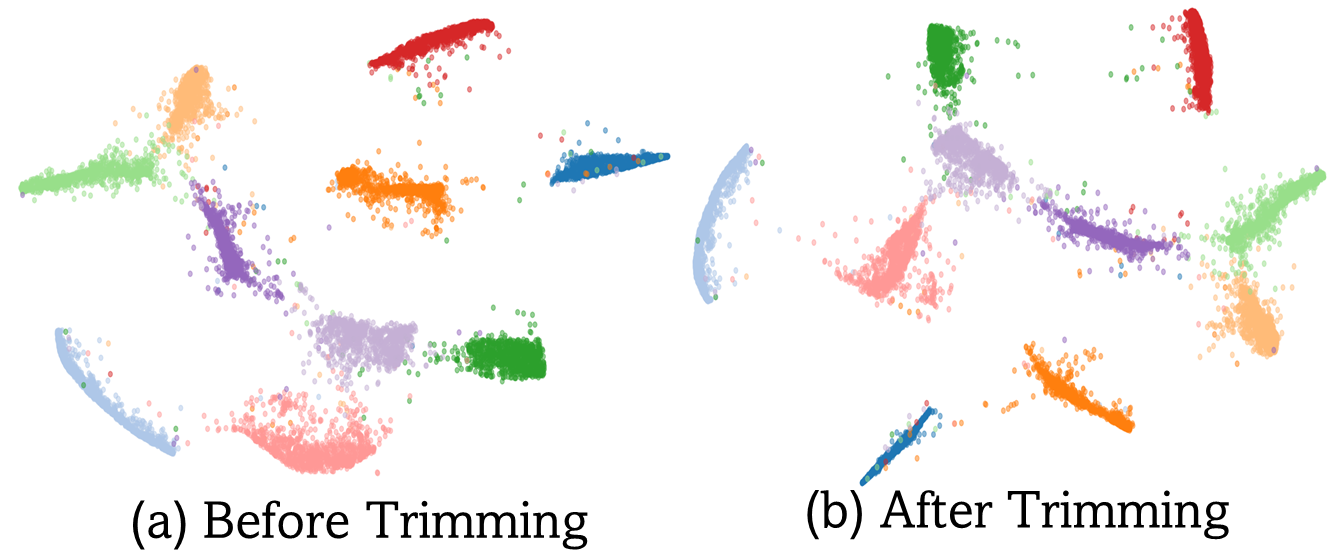}
\caption{The t-SNE visualization of feature embeddings before and after trimming for our proposed trimming approach. It can be demonstrated that the discriminative feature representations can be well maintained, which further demonstrates the effectiveness of our proposed trimming approach. }
\label{Figure_tsne}
\vspace{-0.28 mm}
\end{figure}


\subsection{Benchmark Results}

We did extensive real-world experiments on the public benchmarks to demonstrate the effectiveness and efficiency of our proposed approach. First, as demonstrated in Table \ref{table_youtube}, our proposed approach demonstrates superior accuracy and efficiency while deployed for the video segmentation on the Youtube-VIS benchmark~\cite{yang2019video}. Note that the networks such as MobileNet-V3~\cite{howard2019searching}, Efficient-Net~\cite{tan2019efficientnet}, YOLO-ACT~\cite{yolactpami2022}, and YOLO-Edge~\cite{liang2022edge} are trained with a fully supervised manner, while our proposed approach is tested in an unsupervised manner without leveraging the labeled training data in Youtube-VIS~\cite{yang2019video}. The experimental results demonstrate that our proposed approach realizes even slightly better accuracy in an unsupervised manner compared with the fully supervised counterparts, which demonstrates the effectiveness of our proposed vision-language pre-training approach. Moreover, the efficiency of our proposed approach is much better compared with previous ones and can realize more than 10 Hz, which fulfills the requirements for real-time in diverse robotic applications. 

\begin{figure}[tbp!]
\setlength{\abovecaptionskip}{-0.1cm}
\setlength{\belowcaptionskip}{-0.1cm}
\centering
\includegraphics[width=\linewidth]{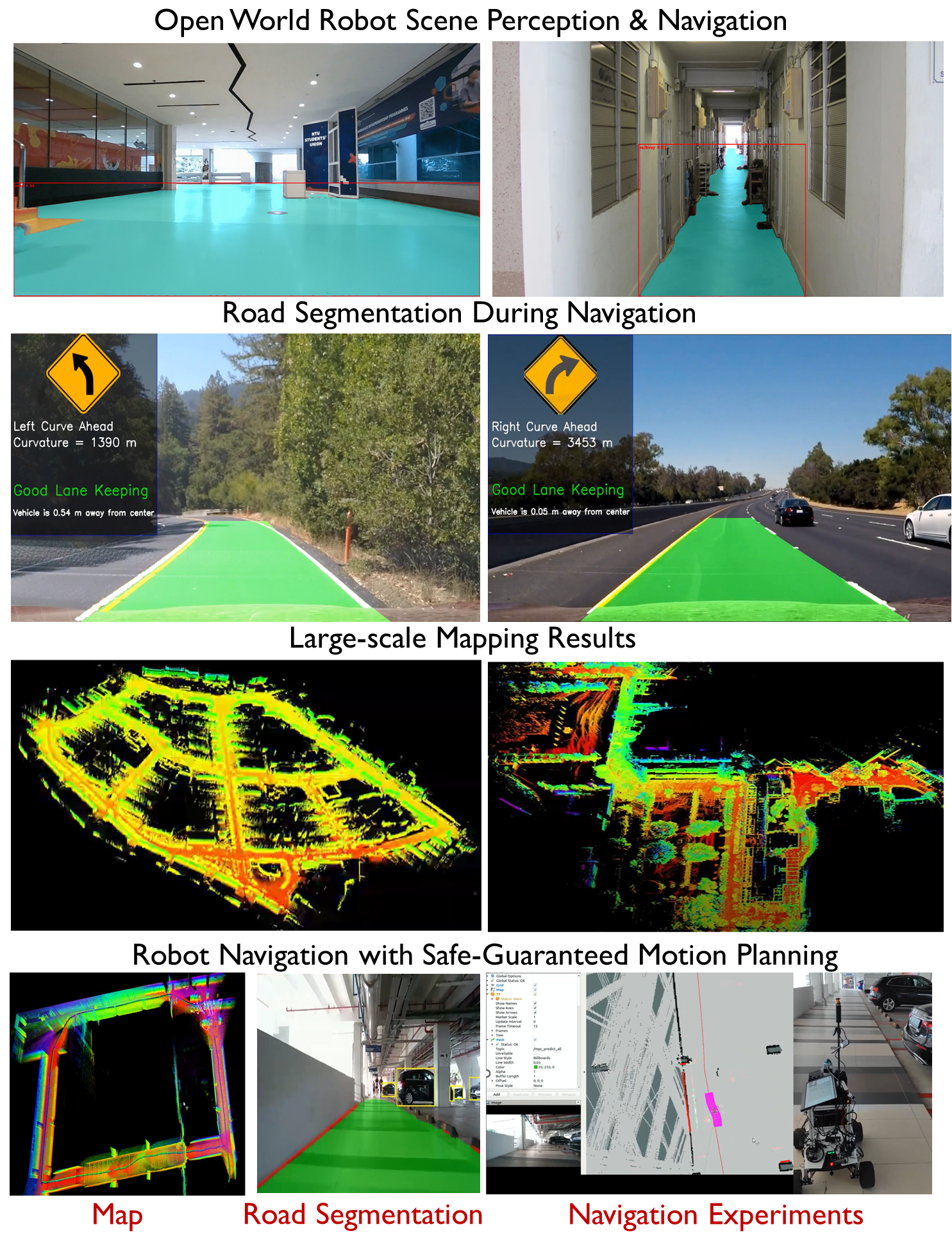}
\caption{The autonomous navigation experiments in real-world situations. It can be demonstrated that our proposed approach can provide accurate segmentation results of the free space of the road and maintain real-time efficiency in the meantime. }
\label{robot_nav}
\end{figure}

\subsection{The Real-World Robot Navigation Experimental Results}
The efficiency of our proposed approach is also validated with real-world experiments as demonstrated in Table \ref{table_ros}. It can be demonstrated that our proposed approach can realize real-time efficiency when deployed onto onboard devices such as TX2, Orin, and Xavier, thus validating the effectiveness and efficiency of our trimming approach.  Also, as demonstrated in Fig.~\ref{Figure_real_world} and Fig.~\ref{building_seg}, the unseen novel classes can be recognized effectively with superb recognition accuracy as demonstrated quantitatively indicated by language prompts. Besides the recognition of the free space on the road, the recognition of other objects such as parking barriers, safe barriers, slogans, and doors are also of significance the language-guided semantic navigation. Also, as shown in Fig. \ref{Figure_tsne}, our proposed approach can maintain well-separated feature embedding space even after pruning, which is in accordance with the quantitative results in Table \ref{table_youtube} that the performance has merely dropped from Mask AP of 62.5\% to 55.8\%. The results both demonstrate discriminative embeddings are learned both qualitatively and quantitatively.
      

Based on our proposed vision-language models and the knowledge distillation strategy, the next step is to integrate the proposed approach into the ground robot system to achieve fully autonomous navigation, which enables the robot to perform autonomous navigation in a robust manner. We conducted extensive experiments on the campus in real-world scenarios as demonstrated in Fig.~\ref{robot_nav}. We adopt the Livox-mid-$360 ^{\circ}$ as our main LiDAR sensor for conducting robot navigation. The robot is required to perform the autonomous navigation task for cargo delivery in a complex human-dense environment. To guarantee safe and collision-free navigation, we are required to provide reliable 3D-free space on the road with high traversability. After obtaining 2D free space, we obtain 3D free space using simple 2D-3D transformation. The 2D-3D transformation can  be done according to the camera intrinsic and extrinsic determining the transformation matrix $\textbf{T}$:
$[\textbf{v}, d] = \textbf{T} [\textbf{p}, 1]$, where \textbf{p} is the 3D point and $\textbf{v}=(u, v)$ is the corresponding pixel location.  It can be demonstrated by our extensive experiments that our system can fulfill tasks of autonomous navigation and can deal with diverse environmental uncertainties in a very effective manner. For the robot scenarios as shown in Fig.~\ref{Figure_system},  we added a safety mechanism that can ensure the 3D position of the robot keeps a safe distance to the road boundary, thus safety can be well guaranteed. As shown in Fig. \ref{robot_nav}, we did extensive experiments in the narrow corridor environment for goal point autonomous navigation. It can be validated that an accurate global map can be obtained and road recognition can be beneficial to finding free space and enable safe-guaranteed robust local motion planning~\cite{liu2022d}. 
Integrating the above modules as a whole, autonomous navigation can be achieved.
    



\section{Conclusion}

In conclusion, we propose an effective framework that deploys current vision-language models to online real-world robot navigation with satisfactory accuracy and real-time performance. On the one hand, we propose a regional language-matching strategy that can effectively enable open-world recognition. On the other hand, we propose the distillation and trimming approach for deploying large-scale vision language models for lightweight real-world robot scene perception and navigation. Extensive experiments demonstrate the effectiveness and efficiency of our proposed approaches. 

\vspace{-1mm}

 \vspace{-0.02cm}

\addtolength{\textheight}{0cm}   





\bibliographystyle{IEEEtran}
\bibliography{references}

\end{document}